# PoseGaze-AHP: A Knowledge-Based 3D Dataset for AI-Driven Ocular and Postural Diagnosis

Saja Al-Dabet[1]*, Sherzod Turaev[1], Nazar Zaki[1], Arif O. Khan[2,3], Luai Eldweik[2,3]

[1] College of Information Technology, United Arab Emirates University, Al Ain, United Arab Emirates
[2] Eye Institute, Cleveland Clinic Abu Dhabi, Abu Dhabi, United Arab Emirates
[3] Cleveland Clinic Lerner College of Medicine of Case Western Reserve University, Cleveland, Ohio, USA

700039885@uaeu.ac.ae,  sherzod@uaeu.ac.ae , nzaki@uaeu.ac.ae,
arif.khan@mssm.edu, eldweil@clevelandclinicabudhabi.ae

## Abstract

Diagnosing ocular-induced abnormal head posture (AHP) requires a comprehensive analysis of both head pose and ocular movements. However, existing datasets focus on these aspects separately, limiting the development of integrated diagnostic approaches and restricting AI-driven advancements in AHP analysis. To address this gap, we introduce PoseGaze-AHP, a novel 3D dataset that synchronously captures head pose and gaze movement information for ocular-induced AHP assessment. Structured clinical data were extracted from medical literature using large language models (LLMs) through an iterative process with the Claude 3.5 Sonnet model, combining stepwise, hierarchical, and complex prompting strategies. The extracted records were systematically imputed and transformed into 3D representations using the Neural Head Avatar (NHA) framework. The dataset includes 7,920 images generated from two head textures, covering a broad spectrum of ocular conditions. The extraction method achieved an overall accuracy of 91.92%, demonstrating its reliability for clinical dataset construction. PoseGaze-AHP is the first publicly available resource tailored for AI-driven ocular-induced AHP diagnosis, supporting the development of accurate and privacy-compliant diagnostic tools.

## 1. Summary and Background

Abnormal Abnormal head posture (AHP) is defined as a deviation of the head from its normal position, characterized by an angular displacement relative to the body [1,2]. It is considered a manifestation of a patient's body language resulting from various factors, including ocular, orthopedic, neurological, or skeletal conditions. The ocular-induced AHPs commonly manifest as head tilt, face turn, chin-up, chin-down, or a combination of these postures. These are often compensatory mechanisms patients adopt to enhance or maintain binocular vision.  Beyond its clinical implications, head movements and postures play a crucial role in nonverbal communication and social interaction. Individuals naturally turn their heads toward others during conversations, nod to convey understanding or agreement, and use additional gestures to express disagreement or contemplation [3,4]. However, when these movements become abnormal, precise diagnosis is essential, as prolonged AHP can negatively impact a patient's quality of life by increasing the risk of falls, dizziness, and difficulties in social interactions. It can also lead to neck strain, pain, and an aversion to visually-oriented tasks [5]. Moreover, AHP may place abnormal stress on the cervical vertebrae, potentially resulting in muscle fatigue, chronic neck pain, or long-term spinal complications [6,7].

Artificial intelligence (AI) has shown remarkable progress in ophthalmology, with deep learning algorithms achieving 90-97% sensitivity in disease detection [8]. However, AI-powered diagnostics require large, diverse datasets that comprehensively capture the specific conditions being diagnosed. Although both observational and instrument-based techniques are used to assess AHP [9–11], accurate diagnosis requires a combined analysis of head posture and eye movement. Existing datasets address these components separately, limiting the development of integrated AI-driven diagnostic tools. Effective AHP diagnosis requires datasets that integrate synchronized capture of AHPs with their underlying ocular conditions. Moreover, access to clinical data is constrained by privacy restrictions, ethical limitations, and the rarity of some conditions, making it challenging to build publicly available, diverse datasets [12].

To address this limitation, we introduce PoseGaze-AHP, the first publicly available 3D dataset that synchronously captures head pose and gaze movement information for ocular-induced AHP analysis. Leveraging peer-reviewed medical literature provides a reliable foundation for constructing knowledge-based datasets while addressing privacy constraints inherent in clinical data sharing. The dataset was created using a novel LLM-based iterative extraction approach with Claude 3.5 Sonnet to systematically extract structured clinical information from 148 medical research papers. Extracted data underwent systematic imputation procedures and was transformed into 3D representations using the Neural Head Avatar (NHA) [13] framework, resulting in 7,920 images representing diverse ocular conditions. This approach preserves patient privacy while providing an authentic clinical data resource that enables the development of more accurate AI-driven diagnostic tools for ocular-induced AHP.

## 2. Method

This section describes the proposed framework, structured into four key phases. The first phase, Data Preprocessing, involves extracting relevant content from ocular research papers. The second phase, Iterative Extraction, utilizes an LLM with structured prompting and feedback mechanisms to refine the extracted data. The third phase, Data Post-processing, focuses on standardizing and imputing the extracted data using medical imputation rules. Finally, the fourth phase, 3D Data Generation, transforms the processed data into 3D head pose and gaze representations. Figure 1 provides an overview of the framework to create the PoseGaze-AHP dataset.

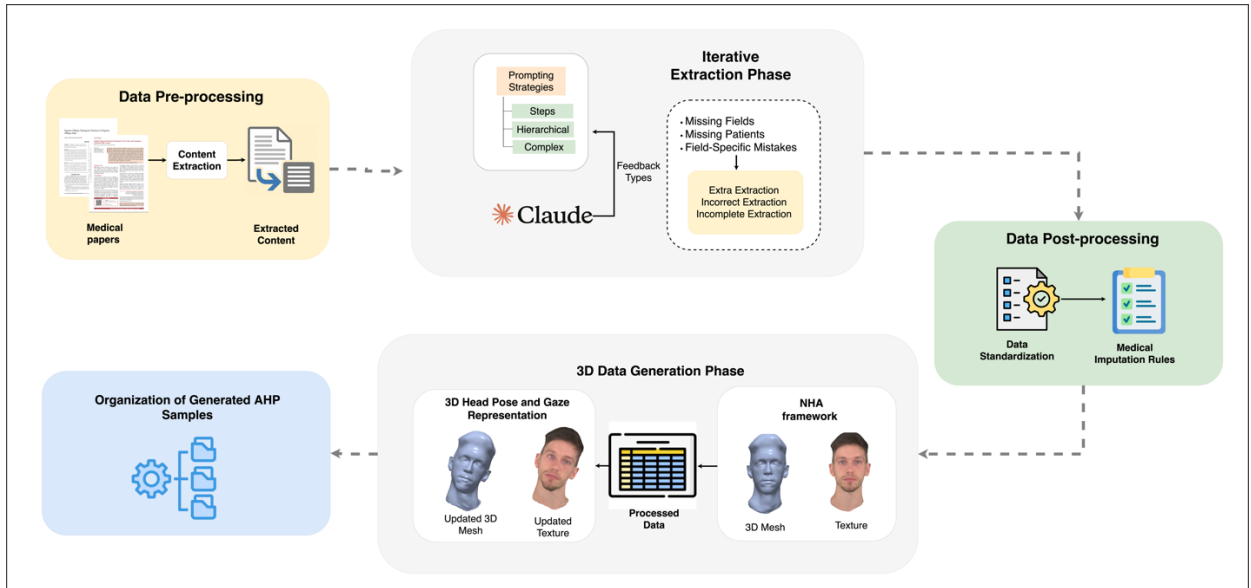

Figure 1: Overview of the framework for creating the PoseGaze-AHP dataset

## 2.1 Ocular Research Papers Collection

A comprehensive set of medical research papers is carefully collected to simulate realistic data for ocular-induced AHPs. This process was initially conducted in our previous systematic review [9], where 750 published medical studies on ocular conditions and their effect on head posture were gathered. A selection process is conducted to narrow the studies down to 180 relevant research works, including case studies, case series, retrospective studies, prospective studies, and review articles. To further refine the dataset, these papers are filtered to include only those containing the necessary information for dataset construction, resulting in a final selection of 148 papers. Based on the papers studied in the systematic review, a set of associations and patterns between ocular conditions and AHPs are extracted. These conditions include Duane Retraction Syndrome (DRS), Nystagmus, Brown Syndrome, Dissociated Vertical Deviation (DVD), Inferior Oblique Palsy (IOP), Monocular Elevation Deficiency (MED), Superior Oblique Palsy (SOP), and other ocular conditions.

The distribution of research papers by methodology and year demonstrates a gradual increase in publications over time, as shown in **Error! Reference source not found.**. The number of published papers has increased over time, with a noticeable rise after 2015. The most significant number occurred in 2021, with 14 papers, followed by 2019 and 2020, which recorded 9 and 8 papers, respectively. Retrospective studies and case series appear to be the most common methodologies, particularly in the later years. Case studies and prospective studies are presented less often. Although less frequent, review papers contribute to the overall distribution, especially in recent years.

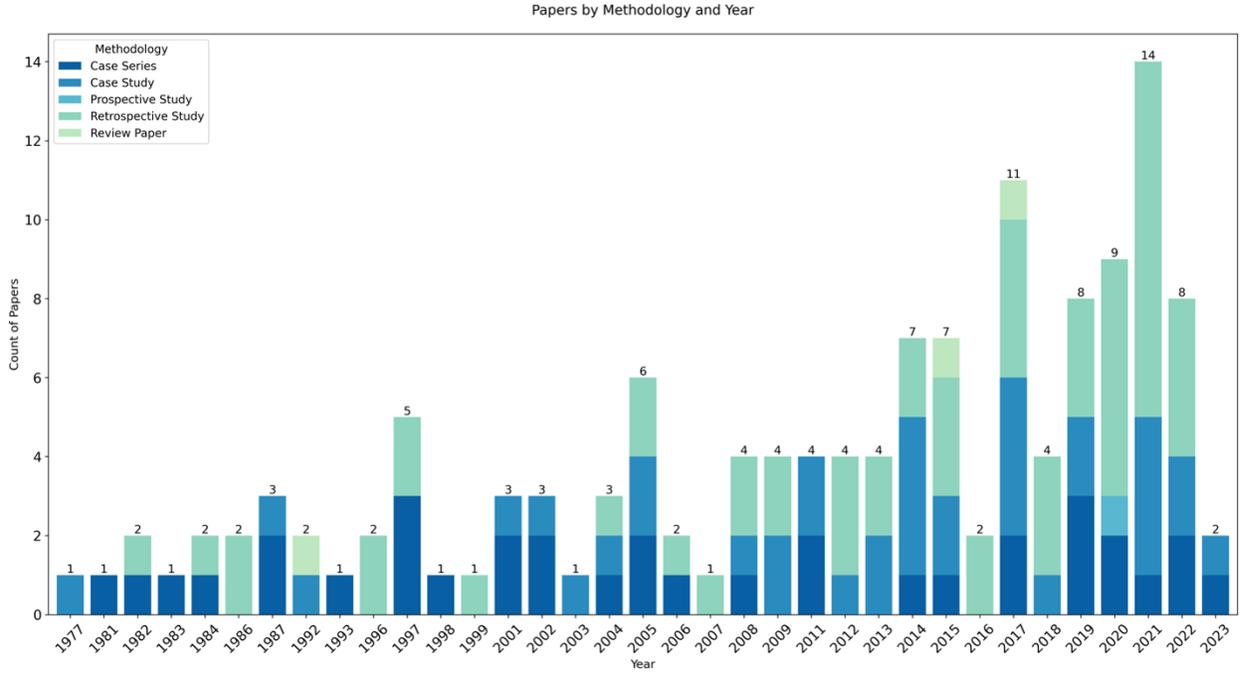

Figure 2: Distribution of Research Papers by Methodology and Publication Year.

## 2.2 Data Extraction Phase

### 2.2.1 Defining key-fields and Preparing Ground Truth

The data extraction phase starts by defining the required information to be obtained from medical papers for dataset construction. This includes key data fields for analyzing ocular conditions and AHPs, which are presented in Table 1. To ensure the accuracy and reliability of the extraction process, a ground truth table is constructed as a reference standard based on the key fields. This table serves as a benchmark for evaluating the correctness and completeness of the extracted data.

**Table 1:** Summary of Key Data Fields for Analyzing Ocular Conditions and AHPs.

| Key Field | Description |
|---|---|
| Paper Title | Title of the research paper |
| Patients No | Total number of patients included in the study |
| Patients Info | Details per individual patients |
| Diagnose | The diagnosed ocular condition |
| Patients AHP No | Number of patients with each type of abnormal head postures |
| AHP Type | Types of abnormal head postures: chin-up, chin-down, head tilt, head turn, or combination of them |
| AHP Direction | Direction of the abnormal head postures: right, left, upward, downward |
| AHP Degree | Degree or angle of the abnormal head postures |
| Eye | The diagnosed eye with the ocular condition left, right, bilateral |
| Eye Misalignment | Type of eye misalignment: hypotropia, hypertropia, esotropia, exotropia |
| PD | Eye misalignment in PD |

*2.2.2 Choosing the LLM Model*

Several LLMs, including GPTs, LLaMA, Gemini, and Claude, can be used for data extraction. While they share similar architectures, they differ in capabilities, accessibility, and specialization. Recent evaluations highlight performance variations across models [14]. GPT-4 excels in factual tasks, achieving high accuracy in question-answering [15], while GPT-3.5 performs well in standard NLP tasks but struggles with complex reasoning [16,17]. In dialogue tasks, both Claude and GPT demonstrate competitive performance [18]. LLaMA-65B is notable among open-source models for its strong deductive reasoning and consistency in complex tasks [19].

Claude, developed by Anthropic, is an advanced LLM that focuses on reliability, transparency, and safety [20]. The Claude 3 series, released in March 2024, includes Claude 3 Opus, optimized for complex analytical tasks; Claude 3.5 Sonnet, which enhances computational efficiency while maintaining high performance; and Claude 3 Haiku, designed for low-latency applications. The recent iteration, Claude 3.5 Sonnet, released in October 2024, exhibits notable improvements in coding, document comprehension, and mathematical reasoning across multiple evaluation benchmarks [21]. Compared to its predecessor, Claude 3.5 Sonnet demonstrates enhanced capabilities in document analysis, chart interpretation, and structured data processing, positioning it as a competitive option among LLMs. Claude 3 Opus offers improved computational efficiency while maintaining high performance, whereas Claude 3 Haiku prioritizes speed but demonstrates lower accuracy in specialized domains such as medicine, law, and finance. These variations show the trade-offs between computational efficiency and domain-specific precision. Given its advancements in multimodal reasoning and structured data interpretation, Claude 3.5 Sonnet is particularly well-suited for applications requiring precise, context-aware analysis, including healthcare and scientific research.

The training framework of Claude 3 models incorporates Constitutional AI and unsupervised learning paradigms, leveraging a diverse dataset comprising publicly available web resources (indexed up to August 2023), licensed third-party datasets, and internally generated data. The development adheres to three core principles: honesty, helpfulness, and harmlessness. Utilizing predictive modeling techniques, the system attains comprehensive language understanding, integrating ethical principles and behavioral guidelines during the reinforcement learning phase, further refined through human feedback mechanisms to enhance reliability [22]. Portions of the human feedback data used in Claude's fine-tuning process have been publicly released [23], along with findings on adversarial testing methodologies and Reinforcement Learning from Human Feedback (RLHF) [24].

Empirical evaluations indicate that Claude 3.5 Sonnet demonstrates superior performance across multiple benchmarks compared to GPT-4 Online (GPT-4o), GPT-4 Turbo (GPT-4T), Gemini 1.5 Pro, and LLaMA 3 400B. Claude 3.5 Sonnet outperforms GPT-4o and Gemini 1.5 Pro in document comprehension tasks. While GPT-4o excels in visual question answering, Claude 3.5 Sonnet exhibits greater proficiency in chart analysis and coding [21]. Additionally, although Gemini 1.5 Pro performs competitively in science diagram interpretation, Claude 3.5 Sonnet maintains greater consistency across all evaluated tasks.

The demand for efficient and accurate data extraction, particularly in the medical research domain, led to the selection of Claude 3.5 Sonnet for extracting key fields from research papers. This phase involves conducting experiments to evaluate a set of LLMs on a sample of research papers, with the results presented in section 4.1.

Claude 3.5 Sonnet was deployed with the following technical parameters: a context window of 1000 tokens, a temperature setting of 0.0 to ensure deterministic outputs and minimize hallucinations, a top_p of 1.0, and a maximum output token of 32,000 per minute. The API calls were managed through a Python-based pipeline using the anthropic package (version 0.49.0), with rate limiting set to respect the model's operational parameters.

*2.2.3 Defining the LLM-based Iterative Extraction Approach*

An effective, prompt engineering process is essential to extract data with specific key fields using LLMs. This involves designing well-structured, task-specific instructions that guide model responses without modifying core parameters. Various prompting techniques are proposed in the literature, such as zero-shot [25] and few-shot [25] prompting for handling new tasks, chain-of-thought (CoT) [26] prompting for enhanced logical reasoning, and retrieval-augmented generation (RAG) [27] to reduce misinformation optimize LLM efficiency and accuracy. Prompt engineering enhances LLM performance by refining user input to generate precise and contextually relevant responses. It consists of instructions (prompts) that provide explicit task guidance and context, which supplies background information to improve comprehension. The LLM, pre-trained on extensive datasets, processes these elements to produce informed outputs. Prompt engineering improves model adaptability by balancing instruction and context and ensuring precise information extraction across diverse domains [28]. This work employs zero-shot and few-shot prompting to extract key fields from medical papers. This approach serves as an initial step in this area, providing clear criteria to guide the model and ensuring precise and reliable data extraction.

Given that the collected medical papers on ocular diseases encompass diverse study types (including case studies, case series, retrospective studies, prospective studies, and review articles), each characterized by distinct structural frameworks and reporting methodologies. Therefore, three prompting strategies are developed for the key field extraction process: step prompt, hierarchical prompt, and complex prompt. These strategies systematically extract and organize data from research papers on ocular diseases and the AHPs mentioned. Each prompt follows a structured approach with defined steps and key fields to ensure comprehensive and standardized data extraction. Steps Prompt: This strategy implements a sequential data extraction methodology, guiding the LLM through three predefined phases: (1) Retrieval of fundamental paper metadata, encompassing title and patient demographics; (2) Systematic categorization of diseases based on diagnoses and AHP distribution patterns; and (3) Extraction of clinical parameters, including AHP characteristics, ocular misalignments, and quantitative measurements. Each phase is augmented with specific location guidance within research papers to ensure systematic and efficient data retrieval. This prompting follows a zero-shot approach, as the strategy is inherently structured in a stepwise manner.

- **Hierarchical Prompt:** This approach implements a tiered extraction strategy, organizing information into three distinct levels of increasing complexity. The framework follows a top-down methodology: (1) Extraction of fundamental study metadata (title, patient demographics); (2) Systematic disease classification and AHP distribution analysis; and (3) Comprehensive documentation of clinical parameters, including ocular misalignments and quantitative measurements. Through this hierarchical segmentation, the strategy facilitates structured cross-study comparisons, which is particularly beneficial for large-scale systematic reviews. This prompting follows a few-shot approach, incorporating a single example to enhance extraction guidance, especially in large-scale reviews.

- **Complex Prompt:** This prompt utilizes a multi-pass extraction strategy, emphasizing precise interpretation of structured clinical data. The framework encompasses three primary components: (1) Systematic analysis of title, abstract, and methods sections for key study parameters; (2) Detailed examination of tabular data for patient demographics, clinical outcomes, and standardized abbreviations; and (3) Comprehensive cross-referencing between extracted data and source text for validation and accuracy verification. This methodology is optimized for studies with extensive quantitative measurements and tabular data presentations. The prompt enforces standardized terminology for AHP and ocular misalignments, ensuring data consistency across diverse research designs. This prompting follows a zero-shot approach, as the prompt illustrates targeted symbols and tabular notations.

presents a comparison of data extraction, and details of the prompts are provided in Appendix A. The three prompting strategies can be described as follows:

- **Steps Prompt**: This strategy implements a sequential data extraction methodology, guiding the LLM through three predefined phases: (1) Retrieval of fundamental paper metadata, encompassing title and patient demographics; (2) Systematic categorization of diseases based on diagnoses and AHP distribution patterns; and (3) Extraction of clinical parameters, including AHP characteristics, ocular misalignments, and quantitative measurements. Each phase is augmented with specific location guidance within research papers to ensure systematic and efficient data retrieval. This prompting follows a zero-shot approach, as the strategy is inherently structured in a stepwise manner.

- **Hierarchical Prompt:** This approach implements a tiered extraction strategy, organizing information into three distinct levels of increasing complexity. The framework follows a top-down methodology: (1) Extraction of fundamental study metadata (title, patient demographics); (2) Systematic disease classification and AHP distribution analysis; and (3) Comprehensive documentation of clinical parameters, including ocular misalignments and quantitative measurements. Through this hierarchical segmentation, the strategy facilitates structured cross-study comparisons, which is particularly beneficial for large-scale systematic reviews. This prompting follows a few-shot approach, incorporating a single example to enhance extraction guidance, especially in large-scale reviews.

- **Complex Prompt:** This prompt utilizes a multi-pass extraction strategy, emphasizing precise interpretation of structured clinical data. The framework encompasses three primary components: (1) Systematic analysis of title, abstract, and methods sections for key study parameters; (2) Detailed examination of tabular data for patient demographics, clinical outcomes, and standardized abbreviations; and (3) Comprehensive cross-referencing between extracted data and source text for validation and accuracy verification. This methodology is optimized for studies with extensive quantitative measurements and tabular data presentations. The prompt enforces standardized terminology for AHP and ocular misalignments, ensuring data consistency across diverse research designs. This prompting follows a zero-shot approach, as the prompt illustrates targeted symbols and tabular notations.

An iterative approach with feedback is utilized to implement the proposed prompting strategies. Each prompt (steps, hierarchical, and complex) is applied iteratively, incorporating feedback at each stage to refine the extracted key fields. The best-performing prompt is then selected, and its results are recorded. The core objective of this approach is to automatically determine the most suitable prompting strategy without manually categorizing research papers, which advances the development of a fully automated extraction process.

**Table 2:** Comparison of Data Extraction Strategies

| Feature | Steps Prompt | Hierarchical Prompt | Complex Prompt |
|---|---|---|---|
| Extraction Methodology | Step-by-step sequence | Tiered categorization | Multi-pass referencing for tables and text |
| Optimal Use Case | Rule-driven, structured extractions | Large-scale reviews with multiple diseases | Studies **with** detailed tables and numerical data |
| Validation Approach | Ensures correctness by following a predefined sequence | Organizes data into hierarchical levels | Ensures numerical accuracy by checking multiple sources |
| Key Advantage | Reduces errors through a structured, rule-driven approach | Facilitates systematic comparison of extracted data across multiple studies | Ensures high precision in extracting and interpreting structured and unstructured data |

The feedback mechanism is designed to enable the LLM to identify mistakes in extracted fields and specify the type of mistake. The process starts by notifying the model of fields that require attention, which serves as initial feedback. Then, three primary mistake categories are defined: missing patient, missing field, and field-specific mistakes. The missing patient category applies to cases where patient information is mentioned in the paper but is not fully extracted by the model (e.g., the paper states that ten patients were studied; however, the LLM extracts data for only six). The missing field category refers to instances where a specific field is explicitly stated for a patient but is not captured by the model. Field-specific mistakes occur when the model extracts the field data but either incompletely, incorrectly, or includes extra values not mentioned in the source paper. These mistakes are identified per each extracted patient record. Table 3 presents samples of feedback type.

**Table 3:** Feedback Samples

| Feedback Type | Feedback Sample |
|---|---|
| Fields Needs Attention | FIELDS REQUIRING IMMEDIATE ATTENTION:<br>- Diagnose<br>- patientsNoAHP<br>- AHPType<br>- AHPDirection<br>- Eye<br>- DegreePD |
| Missing Field | Field 'PD' issues:<br>　No data has been extracted for this field. |
| Missing Patient | NOTE: 18 additional patient records still need to be extracted. |
| Incorrect Extraction | Field 'AHPDegree' issues (Patient-level):<br>  Patient patient_1:<br>  ├── Current: '15'<br>  └── Issue: Are you sure! Numeric value is incorrect |
| Incomplete Extraction | Field 'AHPDirection' issues (Patient-level):<br>  Patient patient_9:<br>  ├── Current: 'left'<br>  └── Issue: Are you sure! Extracted value is incomplete<br>  Patient patient_19:<br>  ├── Current: 'left'<br>  └── Issue: Are you sure! Extracted value is incomplete |
| Extra Values Extraction | Patient patient_3:<br>  ├── Current: '16.5'<br>  └── Issue: Are you sure! Extracted value contains extra information<br>Patient patient_4:<br>  ├── Current: '16.5'<br>  └── Issue: Are you sure! Extracted value contains extra information |

Providing an explicit error type is more effective than generic mistake feedback, which helps the model to better understand extraction mistakes within key fields. Mistakes are identified by comparing extracted data with a ground truth table, where discrepancies can be classified as complete mismatches (e.g., *Eye Misalignment is hypertropia, but the model extracts hypotropia*) or partial mismatches (e.g., *Eye Misalignment is exotropia and hypertropia, but the model only extracts exotropia*). The extracted data from each paper is stored in a JSON file to preserve both paper-based and patient-based key fields.

### 2.2.4 Post-processing Extracted Data

After data extraction, some extracted fields had issues that affected data quality and usability. These include extra text generated during extraction, non-standardized values requiring normalization, and missing data that could lead to incomplete records. A post-processing stage is essential to ensure consistency before advancing to the next phase. The main post-processing steps applied are:

#### A. Data Standardization

This step normalizes the extracted data for each field to ensure consistency before saving and evaluating them in later stages. Standardization is applied based on the nature of each

key field, and the results for all processed papers are stored in a single file. The applied steps include:

- ***Standardizing ocular condition names:*** Standardizing condition names ensures consistency, as researchers use varying terminology, abbreviations, and formats. The process involves converting text to lowercase, removing unnecessary characters (e.g., parentheses, possessives, laterality indicators). A predefined mapping dictionary is then applied to align extracted disease names with their standardized synonyms and variations. For instance, congenital *superior oblique palsy* and *traumatic superior oblique palsy* are standardized under superior oblique palsy, and so on for other synonyms and conditions.

- ***Standardizing AHP Types***: Extracted AHP categories are standardized to ensure consistency. The process begins by converting text to lowercase and removing extraneous spaces. If multiple AHP types are presented, they are separated and processed individually. A predefined mapping dictionary then aligns extracted terms with standardized categories by matching common synonyms and variations. For example, *face turn*, *head turn*, and *face/head turn* are standardized as face turn, while *chin-up* and *chin elevation* are mapped to chin-up, and *chin-down* and *chin depression* are standardized as chin-down. Similarly, head *tilt*, *head tilt*, and *tilt* are unified under head tilt. If an exact match is not found, the extracted term is retained.

- ***Standardizing AHP Direction:*** AHP direction values are standardized to ensure uniformity in terminology. This involves converting text to lowercase and removing extraneous spaces. If directionality is described in relation to the affected eye (e.g., *ipsilateral*, *contralateral*), the corresponding *left* or *right* label is assigned based on the provided eye information. Specifically, *ipsilateral* or *same side* is mapped to the given eye, while *contralateral* or *opposite side* is assigned to the opposite eye (*left* becomes *right* and *right* becomes *left*). For general directional terms, *left* and *right* are retained as is, while vertical directions (*up*, *superior*, *down*, *inferior*) are standardized to *upward* and *downward*, respectively. If no match is found, the original direction term is preserved.

- ***Standardizing Eye Values:*** Variability in eye-related terminology is addressed through a standardization process to ensure consistency in extracted data. As in previous steps, the process begins by converting text to lowercase and removing extraneous spaces. A predefined mapping is then used to standardize different references to the left, right, or both eyes. Specifically, terms such as *OS left eye* and *L* is mapped to the left, while *OD*, *R*, and *right eye* are standardized as right. Similarly, *both eyes* and the *bilateral* are mapped to both in OU. If no exact match is found, the original term is retained.

- ***Standardizing Eye Misalignment:*** Eye misalignment values are standardized by converting text to lowercase, removing extraneous spaces, and standardizing delimiters (e.g., *and*, *with*, + are replaced with commas). Extracted terms are then mapped to standardized categories using a predefined dictionary that accounts for common variations. For example, *HT*, *hyper*, and *hypertropia* are standardized as hypertropia, while *XT*, *exo*, and *exotropia* are mapped to exotropia. Similarly, *ET* and *eso* are unified as esotropia, and *HOT* and *hypo* are standardized under hypotropia.

- **Standardizing Degree Values:** Degree measurements are standardized by unifying formatting, notation, and numerical representation for AHP degree. The process begins by converting text to lowercase, removing extraneous spaces, and eliminating irrelevant terms. A predefined set of rules is then applied to standardize different numerical formats. For instance, *15* is reformatted as 15°, while ranges such as *10-20* or *10 to 20* are converted to 10°-20°. Comma-separated values are processed individually, ensuring all numeric values are correctly formatted with degree symbols.

*B. Handling Missing Data and Preparing Medical Imputation Rules*

To achieve the objective of this study, constructing PoseGaze-AHP dataset for diagnosing ocular conditions and identifying associated AHP, comprehensive data records are required across all relevant fields, including ocular condition, AHP type, AHP direction, AHP degree, diagnosed eye, eye misalignment type, and eye misalignment in PD. However, variability in the types of collected papers (e.g., case reports, retrospective studies, review papers) results in missing values, either due to the nature of the study design or the specific focus of the authors. This poses a significant challenge in generating a complete 3D dataset based on full clinical records. Without a proper data imputation, these data gaps can introduce biases or reduce the dataset's usability in future applications.

Based on our previous systematic review, medical rule-based imputation methods were developed to address this limitation, which analyzed ocular-induced AHPs across 180 studies [9]. These rules provide a structured framework for estimating missing values while maintaining clinical relevance. The imputation process follows a hierarchical approach, starting with the AHP type, which serves as the foundational field for subsequent data completion. Once AHP type is imputed, it can be associated with ocular diagnoses to estimate missing values in other related fields. The process then progresses to the AHP direction, which is calculated based on previously imputed fields. The diagnosed eye is determined thereafter based on the available ocular diagnosis, AHP type, and AHP direction. With these fields established, eye misalignment is estimated, followed by eye misalignment in PD, significantly influencing the AHP degree. In the case of the AHP degree and eye misalignment in PD, average values are used to provide a more balanced and statistically representative estimation, reducing the influence of outliers. By following this structured order, the imputation process ensures logical data consistency. The imputed values are only used to handle missing data and ensure complete records for analysis. A detailed description of these imputation rules is provided in Appendix B.

For example, if the eye misalignment type is missing, it is imputed using a hierarchical approach based on AHP type, diagnosed eye, and paper title. For instance, a study describes a patient with right sixth nerve palsy, a face turn AHP, and a diagnosed eye of right but does not specify the eye misalignment type. The missing value is first estimated by checking for another record within the same paper with the same AHP type and diagnosed eye. If no match is found, and the case is unilateral, the most common ocular condition and AHP type association for unilateral cases are applied – assigning right esotropia, a typical presentation of sixth nerve palsy. If no direct match exists, the most frequently observed AHP type and eye misalignment combination for unilateral cases is used.

## 2.3 Dataset Generation Phase

A set of 3D head modeling techniques can be employed to generate our PoseGaze-AHP. The main requirements include precisely controlling both head pose and eye gaze movements while maintaining a realistic human representation. Among these approaches, the Neural Head Avatar (NHA) framework [13] generates photorealistic 4D head models reconstructed from monocular RGB videos through a hybrid neural architecture. This framework is built upon the Faces Learned with an Articulated Model and Expressions (FLAME) [29] morphable model, which is enhanced through mesh subdivision and further refined by two key neural networks. The Geometry Refinement Network (G) predicts vertex-level offsets to capture detailed facial features and hair structure, while the Texture Network (T) synthesizes view-dependent and expression-dependent textures. The avatar optimization process consists of three stages: initial FLAME model alignment, separate optimization of the geometry and texture networks, and final joint optimization.

The NHA framework provides control overhead pose through a parametric modeling approach that allows yaw rotations (±90°), pitch variations (±15°), and articulated eye gaze movements [13]. To ensure anatomical accuracy during pose transitions, the framework employs advanced geometry refinement techniques, particularly in biomechanically complex regions such as the neck, where pose-induced deformations are most significant. A key advantage of this framework is its ability to maintain geometric consistency while allowing independent control of head pose and gaze. Integrating FLAME's parametric model with neural networks ensures biomechanically realistic poses and generates datasets that capture natural head movements and gaze behaviors [13].

**Table 4:** Camera Configuration Parameters for Multi-View Generation.

| View | Description | Camera Elevation | Camera Side Deviation |
|---|---|---|---|
| Frontal | Standard front-facing view | 0.0 | 0.0 |
| Up | Camera positioned lower, looking upwards | -0.5 | 0.0 |
| Down | Camera positioned higher, looking downwards | 0.5 | 0.0 |
| Left | Camera positioned to the left side | 0.0 | 0.75 |
| Middle Left | Camera slightly shifted left, closer to center | 0.0 | 0.4 |
| Right | Camera positioned to the right side | 0.0 | -0.75 |

| Middle Right | Camera slightly shifted right, closer to center | 0.0 | -0.4 |

After configuring the NHA framework to simulate specific head postures and gaze directions, the imputed dataset is fed into the framework, where key fields are used as generation parameters. All simulated head postures and eye features are presented from the patient's perspective, in accordance with clinical descriptions in the medical literature. For each record, seven distinct views are generated to capture a comprehensive range of head pose variations and to visualize both head posture and eye gaze. Table 4 presents the camera configuration parameters used for multi-view generation. These views include: center, left, right, up, down, middle left, and middle right. The camera is reoriented accordingly: camera elevation (pitch) is adjusted for up and down views, side deviation (yaw) for left and right views, while the camera distance is fixed at 1 for all orientations. Figure 3 presents *(a)* a simulation of head poses with gaze directed forward using the Neural Head Avatar framework, and *(b)* a sample of the multi-view head pose simulations with directional labels defined from the camera perspective.

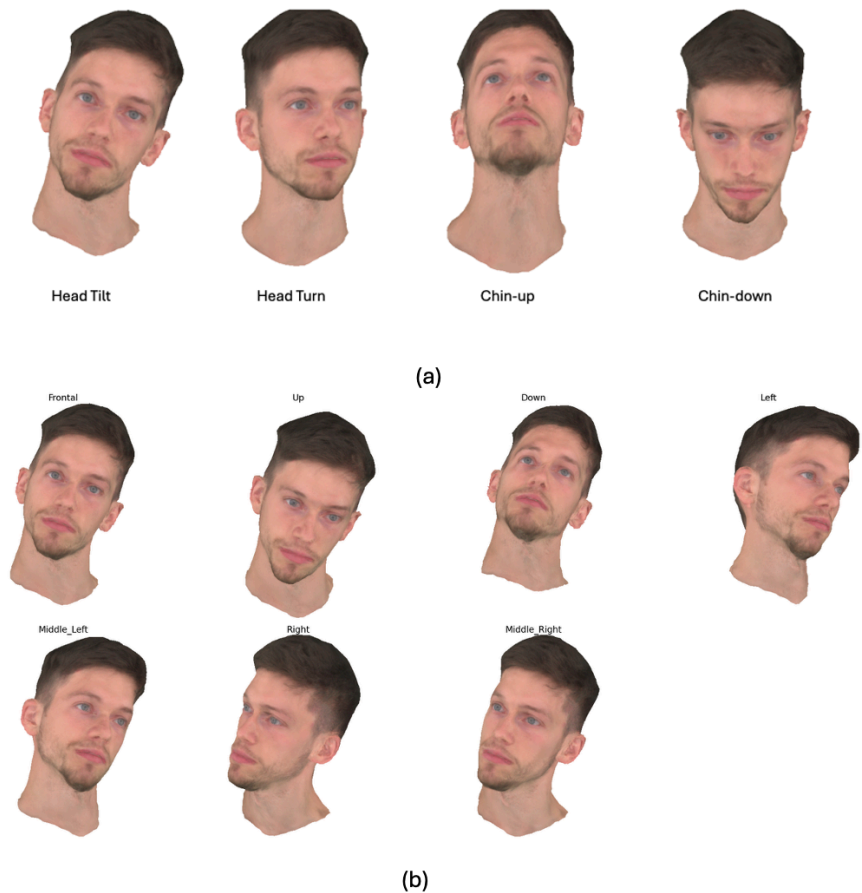

*Figure 3: (a) Simulation of Head Poses Using the NHA Framework with Eye Gaze in Same Direction (b) Simulated Head Poses from Seven Viewpoints Generated Using the NHA Framework*

The generated data is organized with the ocular condition as the primary category and head posture type as a subcategory. Each case is assigned a unique identifier, and within its designated directory, the dataset includes individual head pose images, a combined multi-view representation, and a corresponding metadata file. To ensure transparency, all field values are flagged to indicate whether they have been imputed or not. Two primary head textures, provided by the NHA authors [30], are used for data generation based on their availability.

## 3. Data Records

The PoseGaze-AHP dataset is made available through (https://drive.google.com/drive/folders/1wxRoRmDXElNUgodOkhgpcN9yveNgMki0?usp=share_link). The dataset comprises 495 clinically-derived patient records systematically extracted from 148 peer-reviewed medical research papers, generating 7,920 3D head pose and gaze images for ocular-induced abnormal head posture analysis. The dataset is organized by person (representing different head textures) and individual cases. Each case contains both eye misalignment and AHP data with corresponding viewpoint images. Figure 4 illustrates the dataset hierarchy.

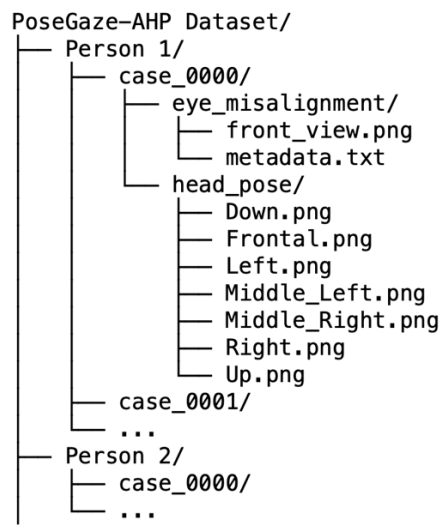

```
PoseGaze-AHP Dataset/
├── Person 1/
│   ├── case_0000/
│   │   ├── eye_misalignment/
│   │   │   ├── front_view.png
│   │   │   └── metadata.txt
│   │   └── head_pose/
│   │       ├── Down.png
│   │       ├── Frontal.png
│   │       ├── Left.png
│   │       ├── Middle_Left.png
│   │       ├── Middle_Right.png
│   │       ├── Right.png
│   │       └── Up.png
│   ├── case_0001/
│   └── ...
├── Person 2/
│   ├── case_0000/
│   └── ...
```

*Figure 4: Folder structure of the PoseGaze-AHP dataset*

Each case in the dataset includes structured visual and clinical components designed to support comprehensive analysis. The dataset consists of PNG-formatted images with standardized dimensions, encompassing seven distinct head pose viewpoints (Frontal, Left, Right, Up, Down, Middle_Left, Middle_Right) alongside frontal eye misalignment representations. Clinical metadata is preserved in structured TXT files containing comprehensive information including ocular condition details, AHP characteristics, eye misalignment parameters, and data imputation indicators. The dual-texture architecture (Person 1 and Person 2) ensures consistent representation across all 495 clinical cases, with each texture generating identical case distributions.

## 4. Technical Validation

This section evaluates key field extraction using Claude 3.5 Sonnet, including model selection, general extraction analysis, and prompting strategies.

### 4.1 Evaluating LLM Model Choice

**Table 5:** LLM Sample Evaluation

| Model | Overall Accuracy | Paper-Level Accuracy | Patients-Level Accuracy |
|---|---|---|---|
| Gemini 1.5 Flash | 85.15 | 93.75 | a.55 |
| GPT 3.5 Turbo | 80.79 | 100 | 61.58 |
| GPT 4o | 86.67 | 95 | 78.35 |
| Claude 3.5 Opus | 86.47 | 100 | 72.95 |
| Claude 3.5 Haiku | 83.70 | 95 | 72.40 |
| **Claude 3.5 Sonnet** | **95.25** | **100** | **90.51** |

Although the primary goal of this paper is not to evaluate LLMs for data extraction, a sample evaluation is conducted to filter out potential options. A random sample of ten research papers was selected to assess the performance of different LLM-based extraction methods. The same data extraction settings are applied across all models for consistency. The evaluation results in Table 5 demonstrate that Claude 3.5 Sonnet outperforms all evaluated models, achieving the highest overall accuracy (95.25%), paper-level accuracy (100%), and patient-level accuracy (90.51%). These results confirm its effectiveness for structured data extraction from research papers and align with previous findings [21] that highlight its advanced document comprehension, multimodal reasoning, and structured data processing capabilities, key factors in its selection for medical research applications. In contrast, GPT-4o (86.67% overall accuracy) and Claude 3.5 Opus (86.47%) also demonstrated strong performance, particularly in paper-level accuracy (95% and 100%, respectively), but their lower patient-level accuracy (78.35% and 72.95%) suggests limitations in extracting granular, structured information. This discrepancy could result from Claude 3.5 Sonnet's refined training methodologies, including Constitutional AI and reinforcement learning techniques, which enhance its ability to interpret complex structured data more reliably. Similarly, Claude 3.5 Haiku (83.70%) and Gemini 1.5 Flash (85.15%) exhibited moderate performance, particularly in patient-level accuracy (72.40% and 76.55%), indicating that these models are optimized for general document comprehension rather than complex analytical reasoning. Their lower scores could be attributed to their design focus on speed and computational efficiency rather than deep contextual analysis, making them more suitable for routine tasks rather than high-precision extractions. GPT-3.5 Turbo had the lowest overall accuracy (80.79%) and significantly lower patient-level accuracy (61.58%), indicating reduced consistency in complex reasoning and structured data processing compared to newer LLMs. This could be due to its less advanced fine-tuning processes and weaker contextual understanding, leading to lower extraction accuracy.

These results justify the selection of Claude 3.5 Sonnet for the targeted medical research data extraction, as it provides better results across different evaluation levels.

## 4.2   General Extraction Analysis

In general, the performance of Claude 3.5 Sonnet on the selected medical papers yielded an overall accuracy of 91.92%, with a paper-level accuracy of 99.31% and a patient-level accuracy of 84.53%. Figure 5**Error! Reference source not found.** presents the field match accuracy for field extraction across 495 records. The results indicate that the model achieves the highest accuracy in extracting paper-level information, such as *"Paper Title"* and *"Patient No",* where these fields are mostly mentioned clearly in the papers. For patient-level accuracy, fields such as *"Diagnosis", "Patient AHP No",* and *"AHP Type"* also exhibit relatively high accuracy. However, certain key-fields, *including "AHP Direction"* and *" Eye Misalignment in PD",* show lower accuracy. For the *"AHP Direction"* field, in many cases, values are not stated in the papers, since authors focus on surgical operations and their effects on eyes, which caused the model to extract incorrect values. For the *" Eye Misalignment in PD"* field, the lower accuracy is reasonable given the challenges in extracting these values from research papers, as they may be presented in different styles and perspectives. Multiple values may be mentioned, or distinctions between preoperative and postoperative values can introduce further complexity, which makes accurate extraction challenging for the model.

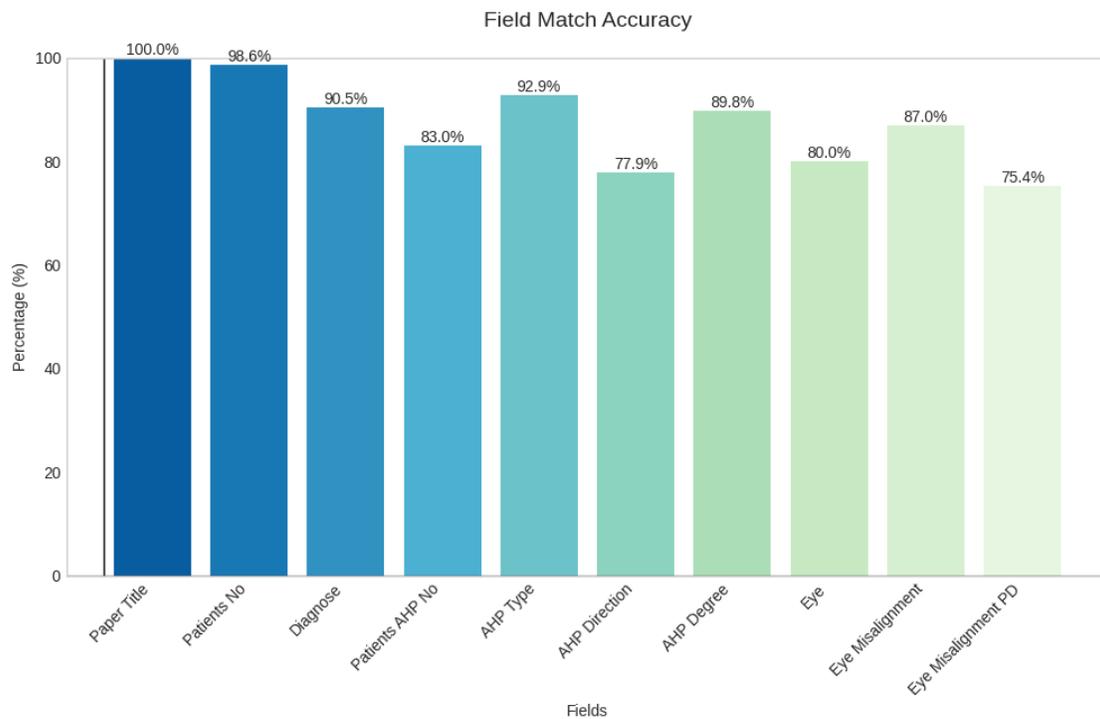

*Figure 5: Overall Field Match Accuracy.*

Figure 6 illustrates the distribution of error types observed in the field extraction process. The most prevalent error category is "Complete Error," accounting for 35.3% of cases where the extracted values were entirely incorrect. *"Missing Patient"* errors constitute 26.5% of cases, indicating instances where patient-related information was not retrieved. In addition, *"Extra Patient"* errors represent 24.0%, suggesting that the model erroneously identifies and extracts patient-related information that is not present in the source paper, a phenomenon commonly referred to as model hallucination. The least frequent error type, *"Partial Match"*, comprised 14.2%, where the extracted values were only partially correct but contained inaccuracies. This indicates that while the model can accurately identify certain portions of the data, it fails to extract the remaining information with complete precision. Although these errors highlight certain limitations in the extraction process, they also indicate the model's ability to recognize and extract relevant information with potential for further refinement and optimization.

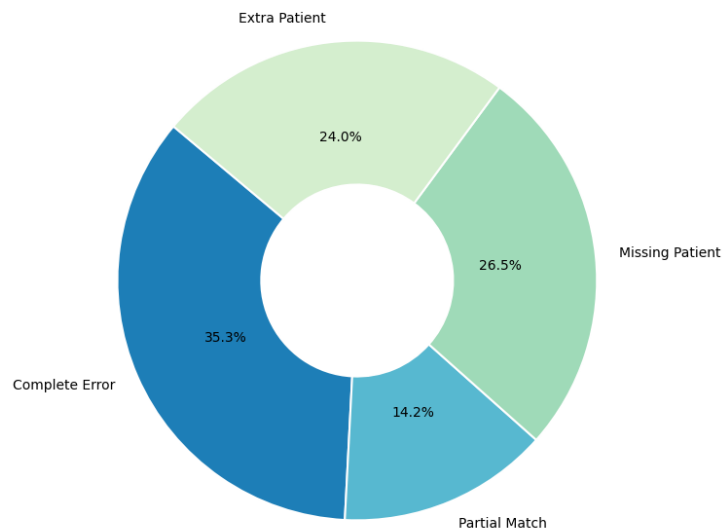

*Figure 6: Error Type Distribution.*

To analyze the error distribution in greater detail, Figure 7 illustrates the field-specific error rates, where both the size and color of the markers represent the error magnitude for each field. Aligning with results in Figure 7, the results show that *"PD," "AHP Direction,"* and *"Eye"* exhibit the highest error rates compared to other fields. Likewise, "Eye Misalignment" and *"AHP Type"* display moderate error rates. In *contrast, "Paper Title"* and *"Diagnosis"* demonstrate minimal error rates.

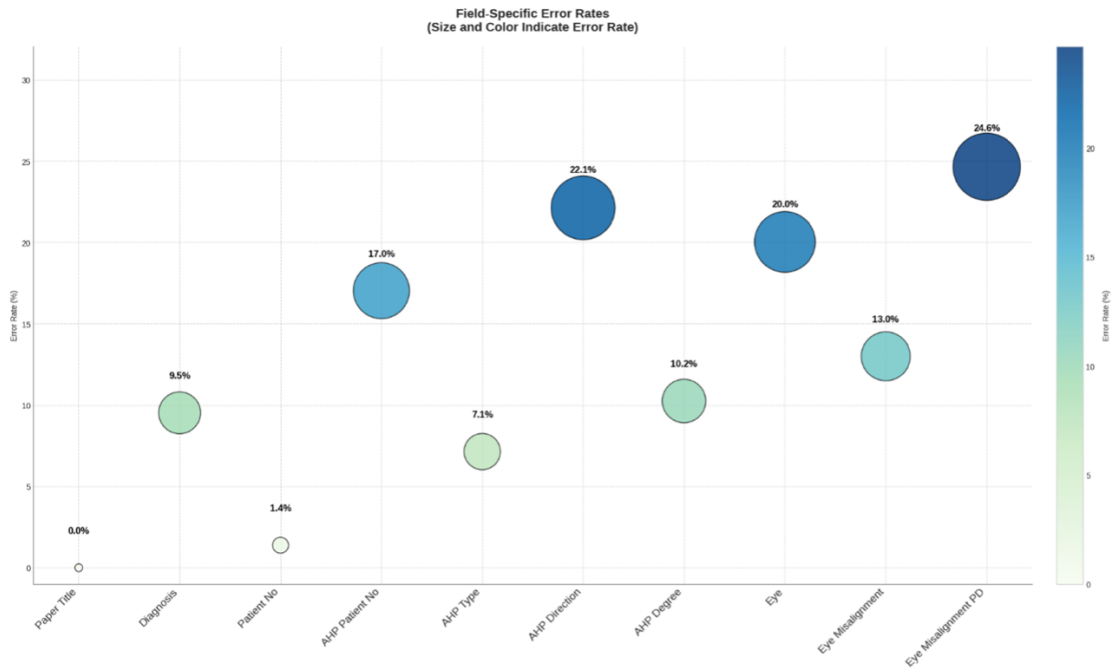

Figure 7: Field-Specific Error Rates.

## 4.3 Prompting Strategy-based analysis

As previously discussed, three primary prompting strategies were developed for data extraction, and Figure 8 presents their respective accuracy rates. The steps strategy achieves the highest accuracy at 92.5%, covering 70.3% of cases, due to its structured, sequential methodology, which systematically guides the model through predefined steps, ensuring efficient data extraction while minimizing errors. The complex strategy, with 90.6% accuracy across 18.6% of cases, employs a multi-pass extraction process, facilitating the interpretation of structured clinical data, particularly in studies with extensive tabular information and quantitative measurements. The hierarchical strategy, achieving 90.2% accuracy in 11.0% of cases, utilizes a tiered extraction framework, supporting systematic cross-study comparisons, especially in large-scale reviews involving multiple diseases. While all three strategies demonstrate high accuracy, the steps strategy is the most effective for structured, stepwise extractions, whereas the complex and hierarchical strategies are better suited for numerical precision and cross-study comparisons.

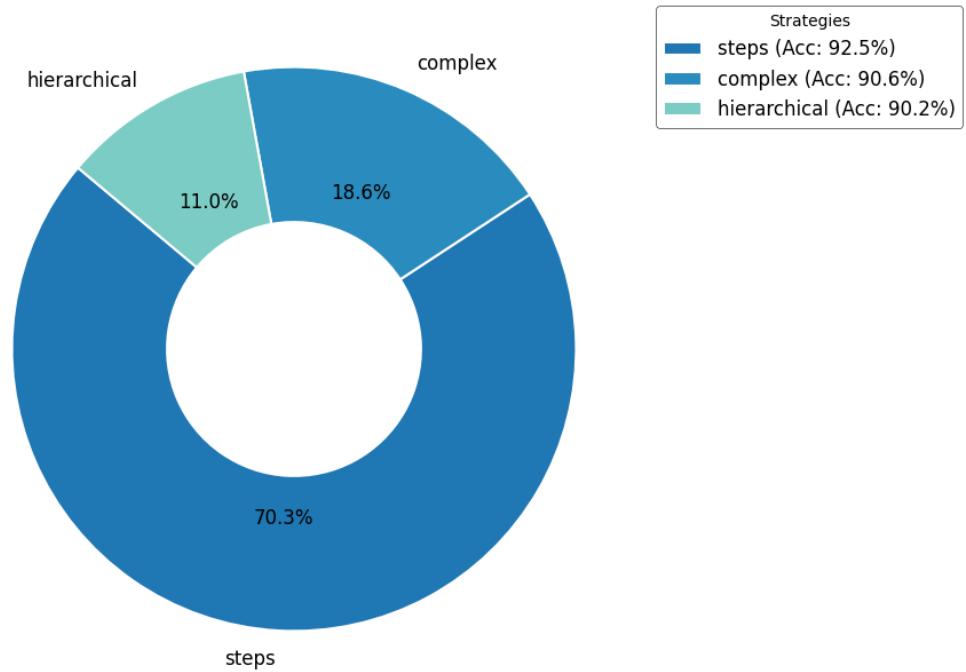

*Figure 8: Strategies Distribution and Performance.*

Figure 9 further examines the distribution of overall accuracy across the three prompting strategies by incorporating a Kernel Density Estimate (KDE) to visualize the probability density of accuracy values. The results indicate that the steps strategy not only achieves the highest accuracy on average but also exhibits the most concentrated distribution near the upper accuracy range, with a peak close to 100%. This pattern aligns with its structured, sequential methodology, which minimizes variability and enhances consistency in extraction performance. In contrast, the hierarchical and complex strategies display broader distributions with lower frequencies with greater variability in their accuracy levels. While both strategies achieve relatively high accuracy, their distributions indicate a more dispersed performance range, particularly in lower accuracy bins. These findings confirm the steps strategy's reliability in general study design extraction while demonstrating the adaptability of the hierarchical and complex strategies in specialized contexts.

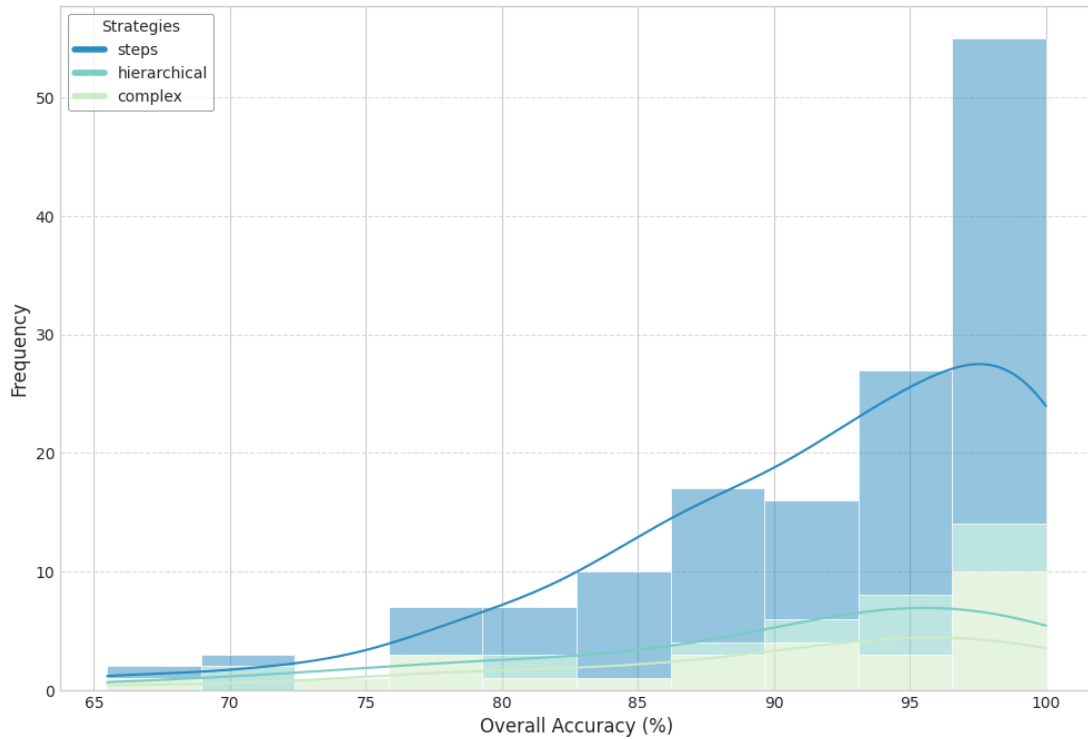

*Figure 9: Overall Accuracy Distribution with KDE by Strategy*

Lastly, Figure 10**Error! Reference source not found.** presents a comparative analysis of accuracy metrics across the three prompting strategies, evaluating overall, paper-level, and patient-level accuracy. The results confirm that the steps strategy consistently achieves the highest accuracy across all metrics, attaining 92.5% overall accuracy, 100.0% paper-level accuracy, and 85.1% patient-level accuracy. This demonstrates its effectiveness in both high-level and detailed extractions. The hierarchical and complex strategies exhibit slightly lower performance, with an overall accuracy of 90.2% and 90.6%, respectively, following similar trends across the paper and patient-level metrics. While both strategies maintain high accuracy, their slightly lower performance in patient-level accuracy (83.6% and 83.0%) indicates greater variability when extracting more granular information.

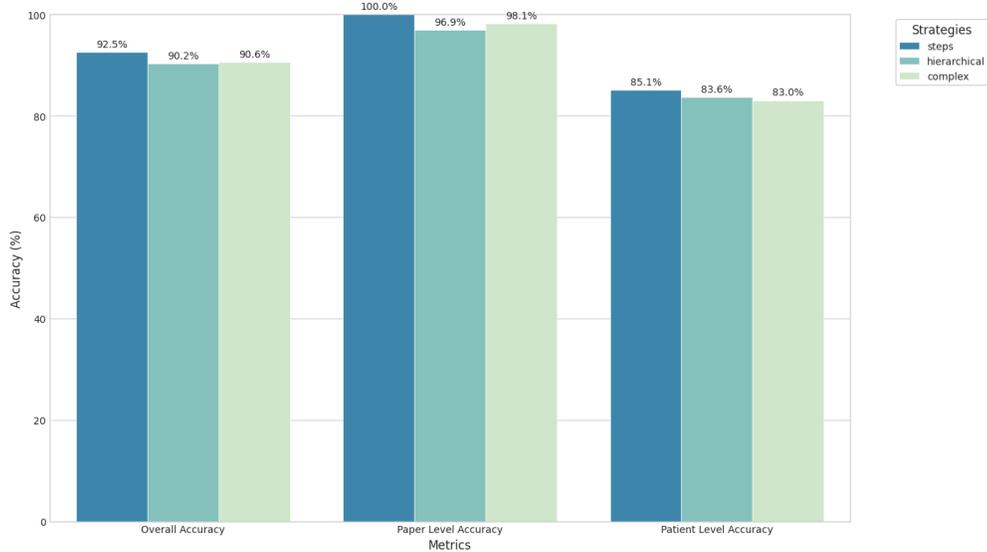

*Figure 10: Accuracy Metrics Comparison by Strategy*

## 4.3 Dataset Description and Statistics

As mentioned earlier, the PoseGaze-AHP dataset is generated based on the ocular conditions covered in the collected research papers. Using Claude 3.5 Sonnet, we extracted 495 distinct records, each corresponding to a patient-group level entry from these papers. A comprehensive statistical summary of the dataset is provided in Appendix C. The dataset exhibits a notable imbalance, primarily due to the distribution of extracted data from the collected studies. This natural imbalance reflects the real-world prevalence of different ocular conditions causing AHP, with Duane Syndrome and Superior Oblique Palsy being significantly more common. While this imbalance accurately represents clinical reality, it poses challenges for AI model training. Future applications using this dataset should consider weighted sampling or other class-balancing techniques to ensure equitable learning across all conditions. However, as this study aims to develop a knowledge-based dataset grounded in real-world ocular medical research, no data augmentation or oversampling techniques have been applied to preserve the integrity and authenticity of the original data. Additionally, two primary head textures were used for data generation, with each texture producing 495 images, resulting in a total of 7,920 images.

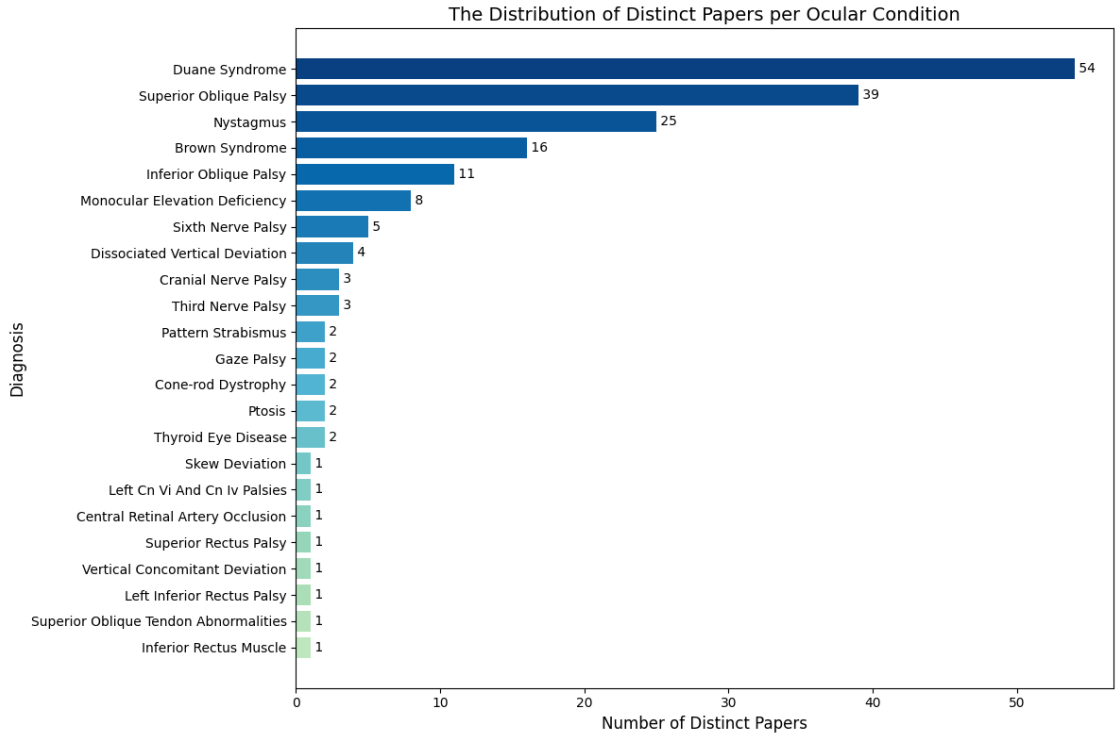

*Figure 11: Distribution of Ocular Conditions in Collected Papers.*

Figure 11**Error! Reference source not found.** illustrates the distribution of ocular conditions associated with AHPs in the analyzed research papers. Duane syndrome (including all types I–IV) and superior oblique palsy were the most frequently reported conditions, documented in 54 and 39 distinct papers, respectively. Nystagmus was the next most common, appearing in 25 papers. Brown syndrome and inferior oblique palsy were also notable, with 16 and 11 papers, respectively. Monocular elevation deficiency was reported in 8 papers, followed by sixth nerve palsy in 5 papers. Dissociated vertical deviation appeared in 4 papers. Less commonly reported conditions included cranial nerve palsy and third nerve palsy (3 papers each), as well as pattern strabismus, ptosis, thyroid eye disease, gaze palsy, and cone-rod dystrophy (2 papers each).

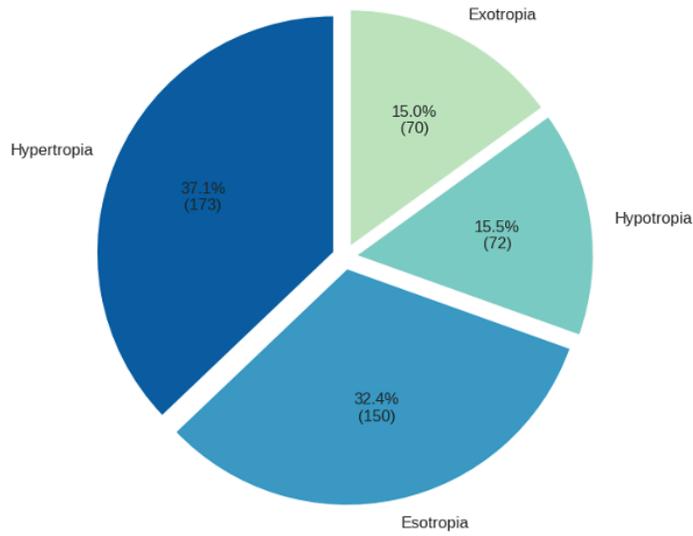

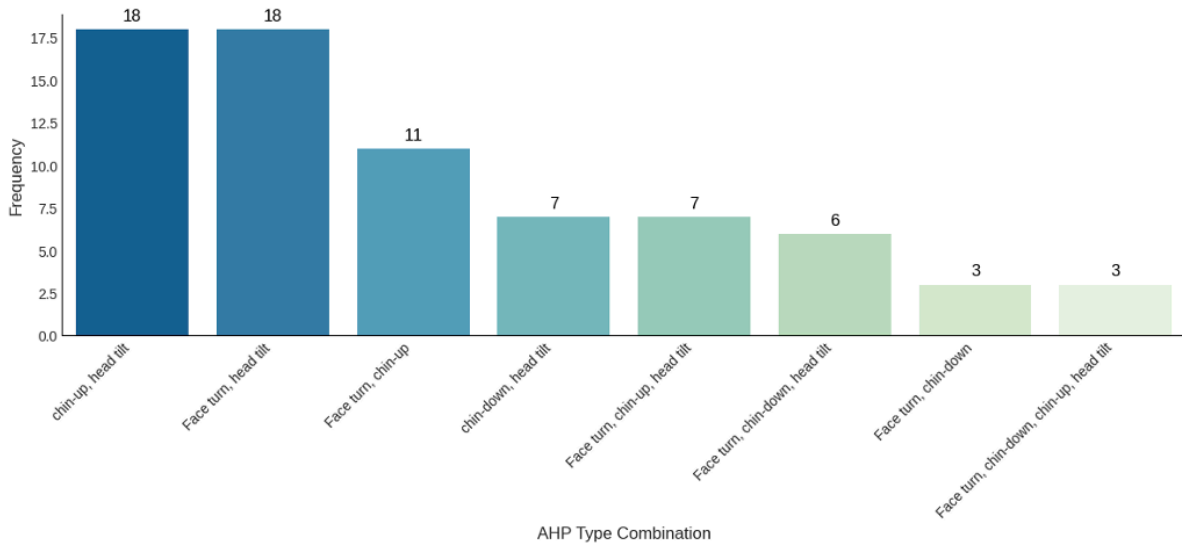

*Figure 12: (a) Distribution of Pure AHP Types (b) Distribution of Combined AHP Types.*

Figure 12 (a) illustrate the distribution of single and combined AHP types while (b) presents the frequency of pure AHP types, showing that face turn is the most common posture (44.0%), followed by head tilt (23.9%). Chin-up and chin-down occur less frequently, at 12.4% and 4.3%, respectively. **Error! Reference source not found.**, a bar chart, depicts the distribution of combined AHP types. The most frequent combinations are chin-up with a head tilt, and face turn with a head tilt, each appearing 18 times, while face turn with chin-up with 11 occurrences. Other combinations appear with lower frequencies, ranging from 7 to 1 instance.

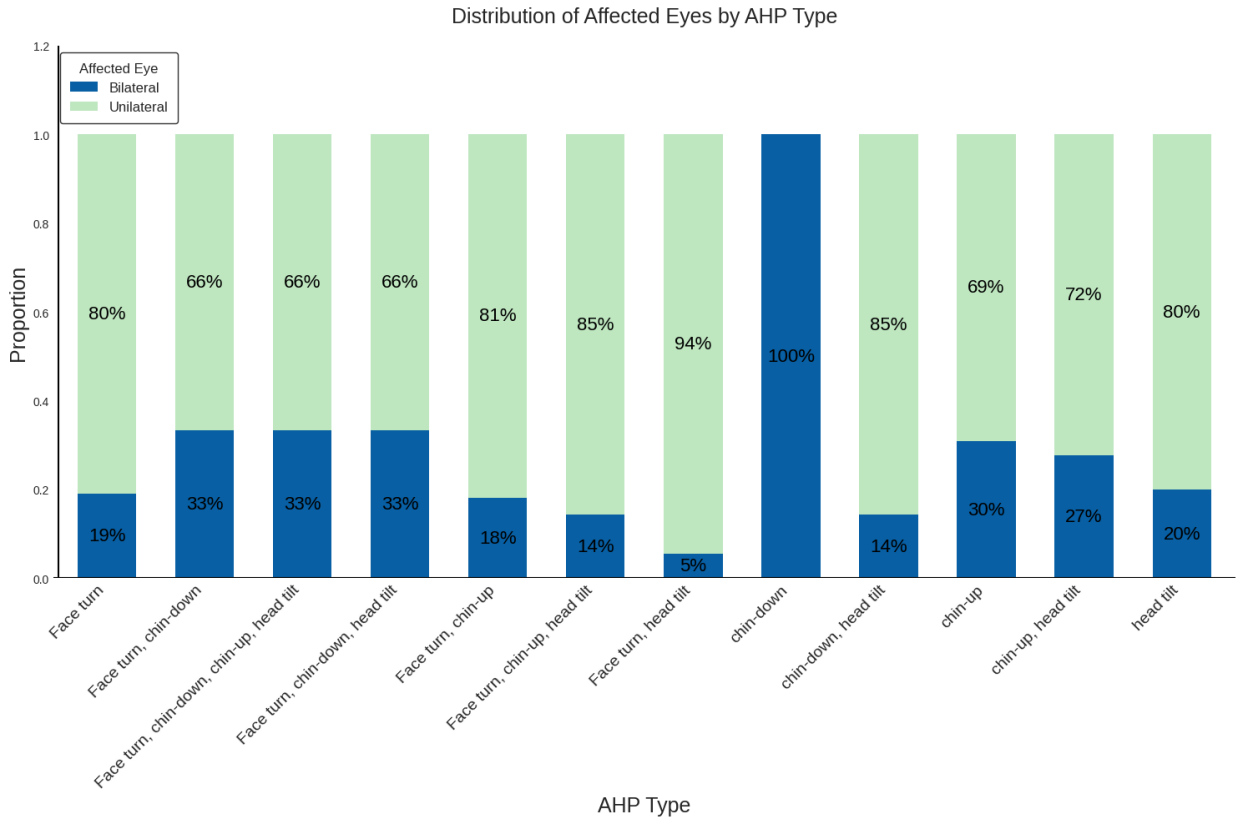

Figure 13: Distribution of the Affected Eyes per AHP Type.

Figure 13 presents the distribution of affected eyes by AHP type, categorized as bilateral or unilateral involvement. The bar chart presents the proportion of each AHP type with respect to eye involvement. In most cases, bilateral involvement is more prevalent, ranging from 60% to 100% across different AHP types. The chin-down posture exhibits 100% bilateral involvement, making it the only AHP type without unilateral cases. Other postures, such as face turn with head tilt (94%) and face turn with chin-up (85%), also show a strong bilateral tendency. Conversely, unilateral involvement is observed more frequently in postures like face turn (33%) and face turn with chin-down (35%).

## Code Availability

All code used for clinical data extraction, structured imputation, and 3D head pose and gaze simulation in the PoseGaze-AHP dataset is publicly available at the following GitHub repository:
https://github.com/saja1994/PoseGaze-AHP-Dataset.git

## Acknowledgments

The authors would like to thank Sandooq Al Watan and the United Arab Emirates University for supporting this work under Sandooq Al Watan Research Grant G00003606 and UAEU Strategic Research Grant G00003676 through the Big Data Analytics Center.

## Funding

This research was funded by the United Arab Emirates University Strategic Research Grant G00003676.

## Author contributions

*Saja Al-Dabet:* Conceptualization, Methodology, Software, Validation, Formal Analysis, Investigation, Data Curation, Visualization, Writing – Original Draft.
*Sherzod Turaev:* Conceptualization, Methodology, Investigation, Resources, Writing – Review & Editing, Supervision, Project Administration, Funding Acquisition.
*Nazar Zaki:* Investigation, Resources, Writing – Review & Editing, Supervision, Project Administration, Funding Acquisition.
*Arif O. Khan:* Validation, Investigation, Writing – Review & Editing.
*Luai Eldweik:* Validation, Investigation, Writing – Review & Editing.

## Competing Interests
The authors declare no competing interests.